\documentclass[a4paper]{article}
\usepackage{booktabs}
\usepackage{bm}
\usepackage{INTERSPEECH2019}

\title{Two-stage Training for Chinese Dialect Recognition}
\name{Zongze Ren, Guofu Yang, Shugong Xu}
\address{
  Shanghai Institute for Advanced Communication and Data Science, Shanghai University, Shanghai, China}
\email{zongzeren@shu.edu.cn}

\begin{document}

\maketitle
\begin{abstract}
In this paper, we present a two-stage language identification (LID) system based on a shallow ResNet14 followed by a simple 2-layer recurrent neural network (RNN) architecture, which was used for Xunfei (iFlyTek) Chinese Dialect Recognition Challenge\footnote{Website: http://challenge.xfyun.cn/aicompetition/global} and won the first place among 110 teams. 
The system trains an acoustic model (AM) firstly with connectionist temporal classification (CTC) to recognize the given phonetic sequence annotation and then train another RNN to classify dialect category by utilizing the intermediate features as inputs from the AM. Compared with a three-stage system we further explore, our results show that the two-stage system can achieve high accuracy for Chinese dialects recognition under both short utterance and long utterance conditions with less training time.

\end{abstract}


\section{Introduction}
The aim of language identification (LID) is to determine the language of an utterance and can be defined as a variable-length sequence classification task on the utterance-level.
The task introduced in this paper is more challenging than general LID tasks cause we use a dialect database which contains 10 dialects in China. The dialects' regions are close to each other and they all belong to Chinese, so they have the same characters and similar pronunciations. 

Recently, the use of deep neural network (DNN) has been explored in LID tasks. The DNN is trained to discriminate individual physical states of a tied-state triphone and then extract the bottleneck features to a back-end system for classification \cite{richardson2015deep,fer2015multilingual,matejka2014neural,jiang2014deep}.
End-to-end frameworks based on DNN later are trained for LID \cite{Lopez2014Automatic}. Other network architectures  are successfully applied to LID task, example for convolutional neural network (CNN) \cite{lozano2015end,Jin2017End}, time delay neural network (TDNN) \cite{garcia2016stacked}, RNN \cite{gonzalez2014automatic,geng2016end,gelly2017spoken}, and \cite{cai2019utterance} has a CNN followed by an RNN structure, which is similar to ours.
They predict the final category of an utterance directly by the last fully connected layer, or derive the results by averaging the the frame-level posteriors.
These frameworks just trained end-to-end to recognize languages, but they do not consider the phonetic information concretely.

On the other hand, in many utterance analyzing tasks such as acoustic speech recognition (ASR), speaker verification (SV) and our LID, only a simple task or a specific aim is focused on.
However, an utterance always has multi-dimensional information such as content, emotion, speaker and language and there are some certain correlations between them.
Although the LID task is text-independent, which means the content of each utterance is totally different, different languages may have its own pronunciations or tones. 
Thus acoustic and language are two components in the LID task, \cite{lozano2018dnn,Tang2017Phonetic,Masumura2017Parallel,Mateju2018} use the bottleneck features from an ASR system and feed to another neural network for recognition.
Nevertheless, these ASR DNNs constituted by fully connected layers adds significant computational complexity and also require labels of physical states of a tied-state triphone.

Inspired by all this, we assume that the high-dim features extracted from the network will contain information of pronunciation and language category, so we combine the ASR method with our LID task and here are our main contributions:

\begin{itemize}
\item[1.] We explore a new Chinese dialect database by a two-stage LID system and achieve high accuracy of 89\%. Our system won the first place in Xunfei (iFlyTek) Chinese Dialect Recognition Challenge, which attracted 110 teams joining in. 
\item[2.] Compared with other methods which recognize physical states or triphones, we train an AM end-to-end and use CTC \cite{graves2006connectionist} to recognize the phoneme sequence of an utterance directly. 
\item[3.] We use a specially designed ResNet14 followed by an RNN instead of using fully connected layers to develop the two-stage LID system. We train an AM in the first stage and then use the intermediate features from the AM as inputs to train an RNN to compute posteriors for LID in the second stage. 

\item[4.]
We further investigate a three-stage system, we first train an AM to align the phoneme label and then train another AM to predict the phoneme of each frame, finally we use the intermediate features from the second-stage's network to train an RNN. The results show that the performance is slightly worse than the two-stage system.
\end{itemize}
The remainder of the paper is organized as follows. Section 2 introduces some related works about ASR and section 3 introduces the ResNet14 structure and gives processing of the two multi-stage systems. We present details of the database and initialization methods of the network in Section 4. The results of experiments and analysis are shown in Section 5. Lastly we give the conclusion and some future work in Section 6.

\section{Related works}
ASR \cite{hinton2012deep} task enables the recognition and translation of spoken language into text. Traditionally, we can train an AM based on frame-wise cross-entropy loss to recognize phoneme, which requires tedious label alignment procedure such as Hidden Markov Model and Gaussian Mixture Model (HMM-GMM) paradigm. Then we can use a pronunciation model and a language model to transfer into text. 

In the latter case, CTC is used in training end-to-end ASR network \cite{graves2013speech,graves2014towards,amodei2016deep}, which means that we do not have to align the phoneme label before training. 
CNN followed by RNN architectures have shown strong ability in dealing sequence-related problems such as sense text recognition \cite{shi2017end} and ASR \cite{zhang2016towards}.
These make the ASR network easy to train and perform better with fewer parameters. \cite{wang2017residual,kim2017residual} further add residual links to the CNN and RNN respectively and both make significant progress.

\section{Two-stage system overview}
\subsection{Network structure}

The major network structure we use in the two-stage system can be divided to the CNN part and the RNN part, as described in Table \ref{crnn_table}. Given the input data of shape $T \times 40$, where $T$ is the frame length of an utterance, we finally get 512-dimensional frame-level representation and $N$ is the number of phonemes or dialect categories. 

Compared with other DNN based systems, we design the CNN part based on ResNet-18 \cite{he2016deep} structure, named ResNet14, as the main part, which decreases the parameters a lot. 
 The first conv layer is with kernel size $k=7\times7$ and stride size $s=2$, followed by a maxpool layer with stride size $s=2$ for downsampling. Then the residual blocks extract high-dim features from the input sequences and keep the low-rank information.
There are 6 res-blocks in all for decreasing parameters, the kernel size of each block is $k=3\times3$ and the features are downsampled while adding channels.

We use 2-layer bidirectional long-short term memory (BLSTM) \cite{schuster1997bidirectional} as the RNN part following ResNet14. BLSTM extends original LSTM by introducing a backward direction layer so it considers the future context. The output of the network will be linked to different loss functions and different labels in different stages.

\renewcommand\arraystretch{1.3}
 \begin{table}
 \caption{Our ResNet14 structure, which has only 5.36 million parameters compared to the standard ResNet-18 (11.26 million)}\label{crnn_table}\centering
 \begin{tabular}{|c|c|c|c|}
 \hline
Layer &  Channels & Blocks &Output size\\
 \hline
Conv1&64&N/A& $\frac{T}{2}$ $\times$  64$\times$20\\
  \hline
 Maxpool&N/A&N/A&$\frac{T}{4}$ $\times$  64$\times$10\\
 \hline
Res Conv1&64&2& $\frac{T}{4}$ $\times$  64$\times$ 5\\
 \hline
Res Conv2&128&2& $\frac{T}{4}$ $\times$  128$\times$ 3\\
 \hline
Res Conv3&256&1& $\frac{T}{4}$ $\times$  256$\times$ 2\\
 \hline
Res Conv4&512&1& $\frac{T}{4}$ $\times$  512$\times$ 1\\
 \hline
BLSTM1&N/A&N/A&$\frac{T}{4}$ $\times$  512\\
 \hline
BLSTM2&N/A&N/A&$\frac{T}{4}$ $\times$  512\\
 \hline
Output&N/A&N/A&$\frac{T}{4}$ $\times$  N\\
 \hline
\end{tabular}
\end{table}
\renewcommand\arraystretch{1.15}
\subsection{Loss function}
CTC is an objective function that allows an RNN to be trained for sequence transcription tasks without requiring any prior alignment between the input and target sequences. The label sequence $z$ can be mapped to its corresponding CTC paths. We denote the set of CTC paths for $z$ as ${\rm p\in\Phi (z)}$. Thus the likelihood of $z$ can be evaluated as a sum of the probabilities of its CTC paths:
\begin{equation}
P{\rm(\bm{z}|\bm{X})}=-\sum_{{\rm p\in\Phi(z)}}^{}P{\rm(\bm{P}|\bm{X})}
\end{equation}
where $X$ is the utterance and $p$ is a CTC path. Then the network can be trained to optimize the CTC function $-ln P{\rm(\bm{z}|\bm{X})}$ by the given sequence labeling.
For the LID task, we use the multi-class cross-entropy loss for classification:
\begin{equation}
L_{CE}=-\sum_{j=1}^{T}y_ilogP_j
\end{equation}
where $y_i$ is the ground truth label and $P_j$ is the output probability distribution.
\subsection{Two-stage system}
\begin{figure}
\centering
\includegraphics[scale=0.52]{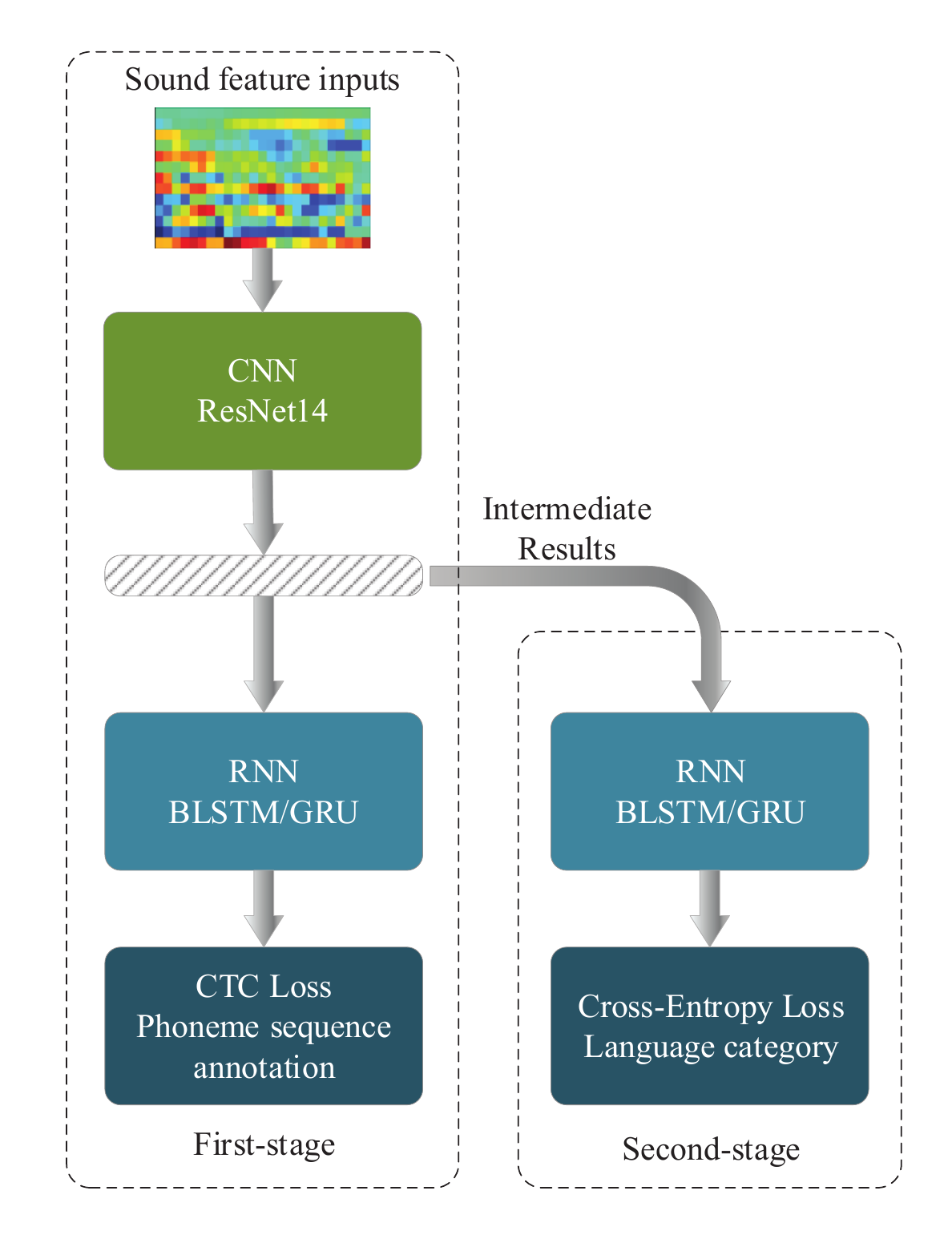}
\caption{Training architecture of two-stage system} \label{fig2-network}
\end{figure}

Figure \ref{fig2-network} shows the architecture of the two-stage system. The input is the sound feature of each utterance. We firstly train the AM with the ResNet14 followed by an RNN architecture, then the intermediate results computed by res-blocks are feed to the second stage as the input. The framework does not need to compress the feature sequence so it keeps all the information from the ResNet14 part. The network of second stage is 2-layer BLSTM. The final pooling strategy is average pooling on time-dimension so we can get the utterance-level category results from frame-level, and the output is the prediction of dialect category. 
We use CTC loss to train the AM so the network outputs can align with the phoneme sequences automatically and use cross-entropy loss to discriminate between dialects.


Compared with multi-task training \cite{Liu2018,zhou2018training} in SV tasks, it should be emphasized that these stages should be trained step by step instead of multi-task learning with shared layers, that is to say we backpropagate the whole network while training AM, and only backpropagate the RNN part in the second stage, or the network will be degenerated and lost the information of acoustic knowledge.

\subsection{Three-stage system}
\begin{figure}
\includegraphics[scale=0.5]{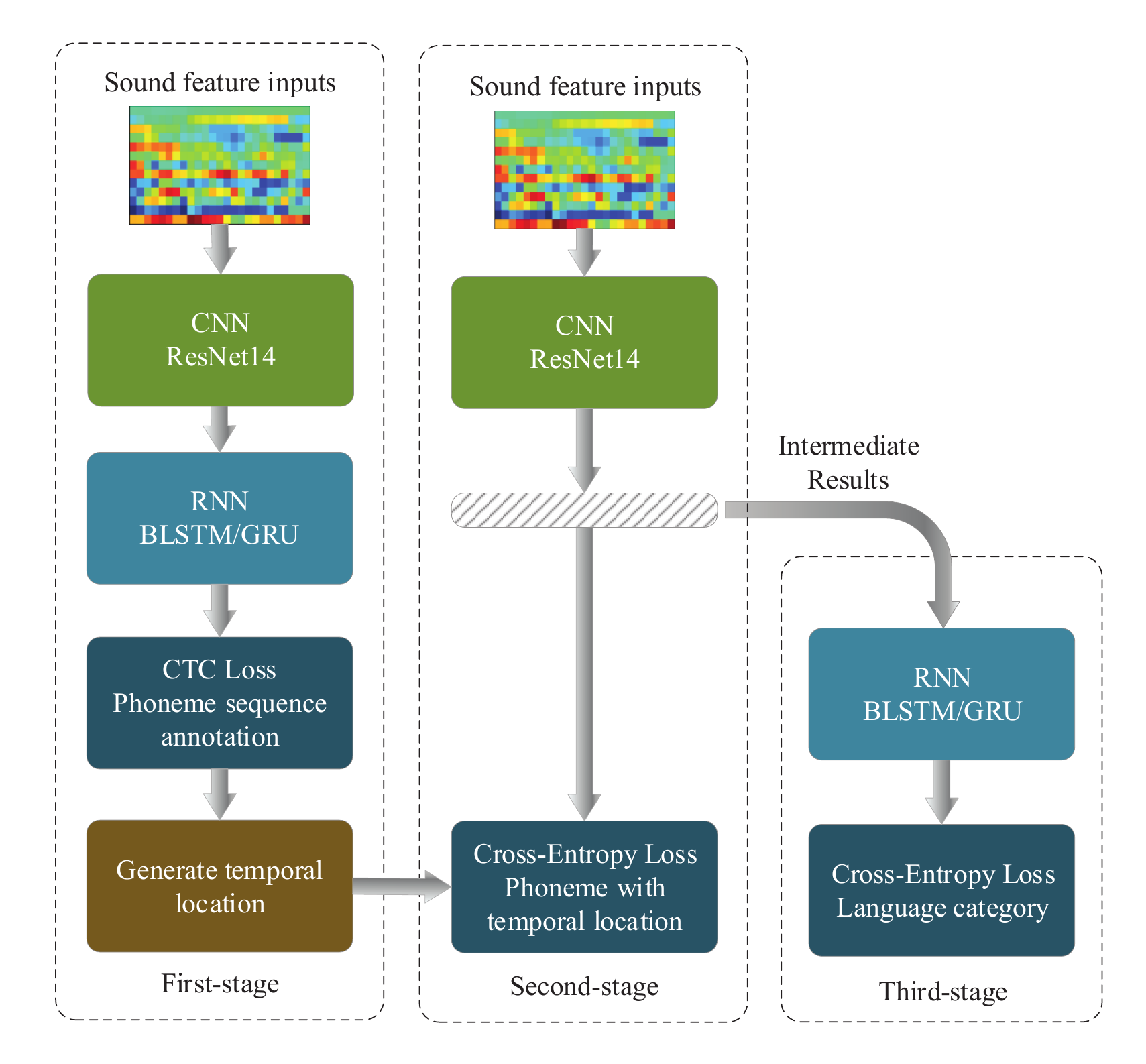}
\caption{Training architecture of three-stage system} \label{fig3-network}
\end{figure}

The three-stage system, as shown in Figure \ref{fig3-network}, has a more complex architecture. Firstly, we still train an AM whose architecture is the same as the first-stage in the two-stage system. This AM is used to generate temporal locations of each phoneme through CTC loss, so that we can train an another AM by using cross-entropy loss as the second-stage to predict the corresponding phonetic labels of the input frames, in which we only use ResNet14 without an RNN because we have the precise locations of each phoneme from the first stage.
The third stage is similar, we use the intermediate features from the second stage to train an RNN network for LID task, also the loss in this stage is cross-entropy loss. 

\section{Experiments}
\subsection{Data description}
\begin{figure}
\includegraphics[scale=0.55]{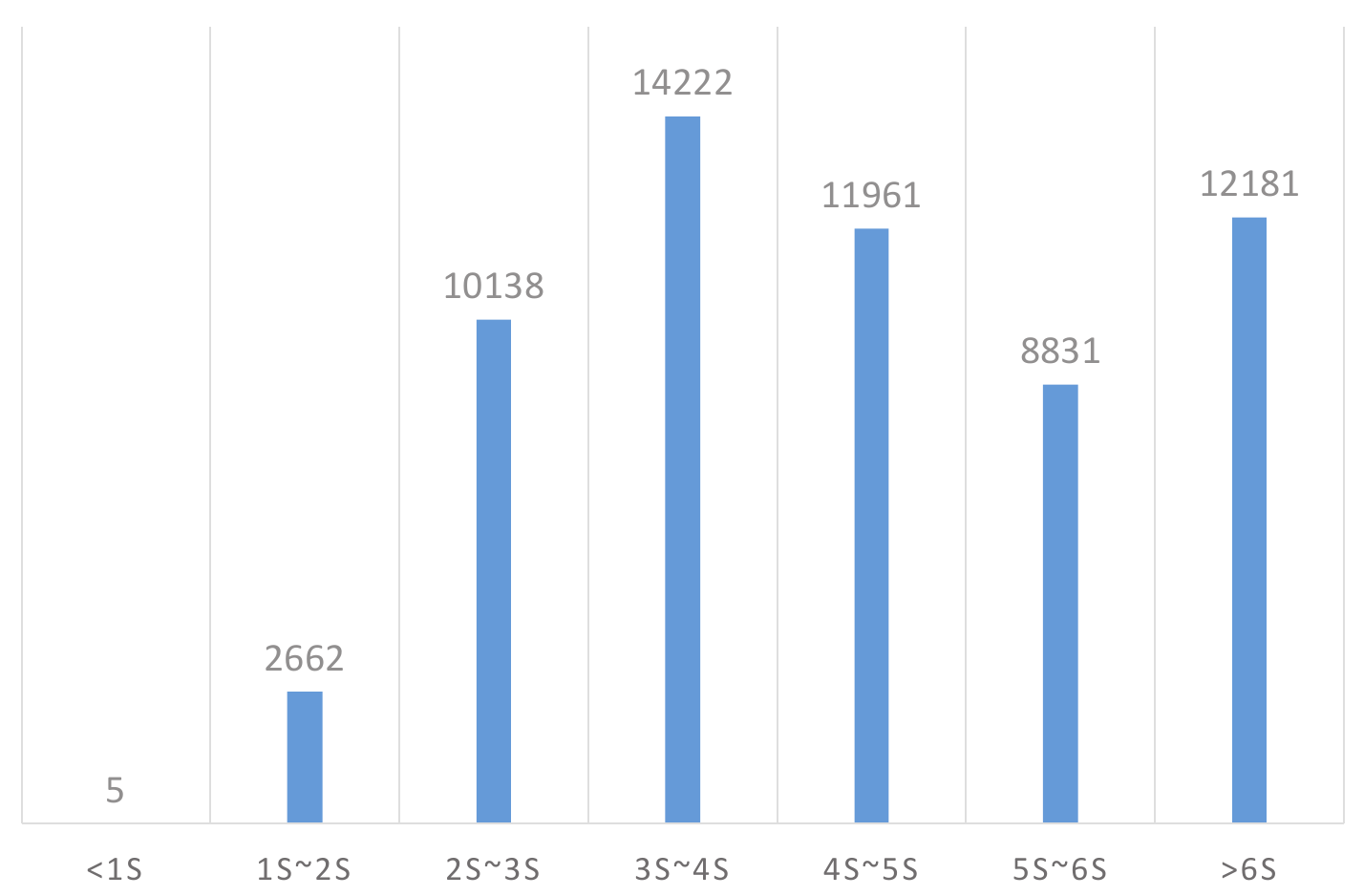}
\caption{Data distribution of time} \label{data}
\end{figure}

We use a database covering 10 most widespread Chinese dialects, the dialects are Ningxia, Hefei, Sichuan, Shanxi, Changsha, Hebei, Nanchang, Shanghai, Kekka and Fujian. Each dialect has 6-hour audio data.
For the training set, there will be 6000 audio files in each dialect with variable length (Figure \ref{data}), we can see that most files are longer than 3 seconds. The test set has 500 audio files in each dialect and the set is divided into two categories according to the duration of the audio file ($\leq$3s for the first task and $>$3s for the second task).

The phonetic sequence annotation of the corresponding text to each speech is also provided in the training set. There are 27 initials and 39 finals with 148 tones in the whole database.
\subsection{Experimental setup}
We convert the raw audio to 40-dimensional log Mel-filterbank coefficients with a frame-length of 25 ms, mean-normalized over the whole utterance. Then we stack all the log-mel filterbank features and feed it into the neural network, which is implemented in PyTorch. No voice activity detection (VAD) or other data augmentation approaches are applied. During the training process, we use Adam as the optimization method and set different learning rates and weight decay in different stages. We do not set dropout while training AM but set the dropout value=0.5 while training the LID network (the last stage). 

The baseline we use for comparison is a one-stage RNN system, the RNN structure is the same as the last stage containing 2-layer BLSTM and directly trained to recognize dialect category. In the process of evaluation, we compute the accuracy of the two sub-tasks and the whole test set to evaluate the performance of each system.

\section{Results}
\subsection{Comparison of different stage systems}
First of all, we compare the two-stage system and the three-stage system trained with phonetic sequence annotation and dialect category label with the baseline trained only with dialect category label. The two multi-stage system have the same ResNet14 architecture and use 2-layer BLSTM as the RNN part with 256 nodes.
From the results in the Table \ref{tab2}, we can see that the relative accuracy (ACC) of the two multi-stage systems increases by 10\% on every task relative to the baseline and the two-stage system performs best. We also observe that both two multi-stage systems perform excellently in long duration ($>$3s) task and the two-stage system illustrates its advantageous and robustness in short duration ($\leq$3s) task.

\begin{figure*}[!t]
\centering
 {\includegraphics[scale=0.58]{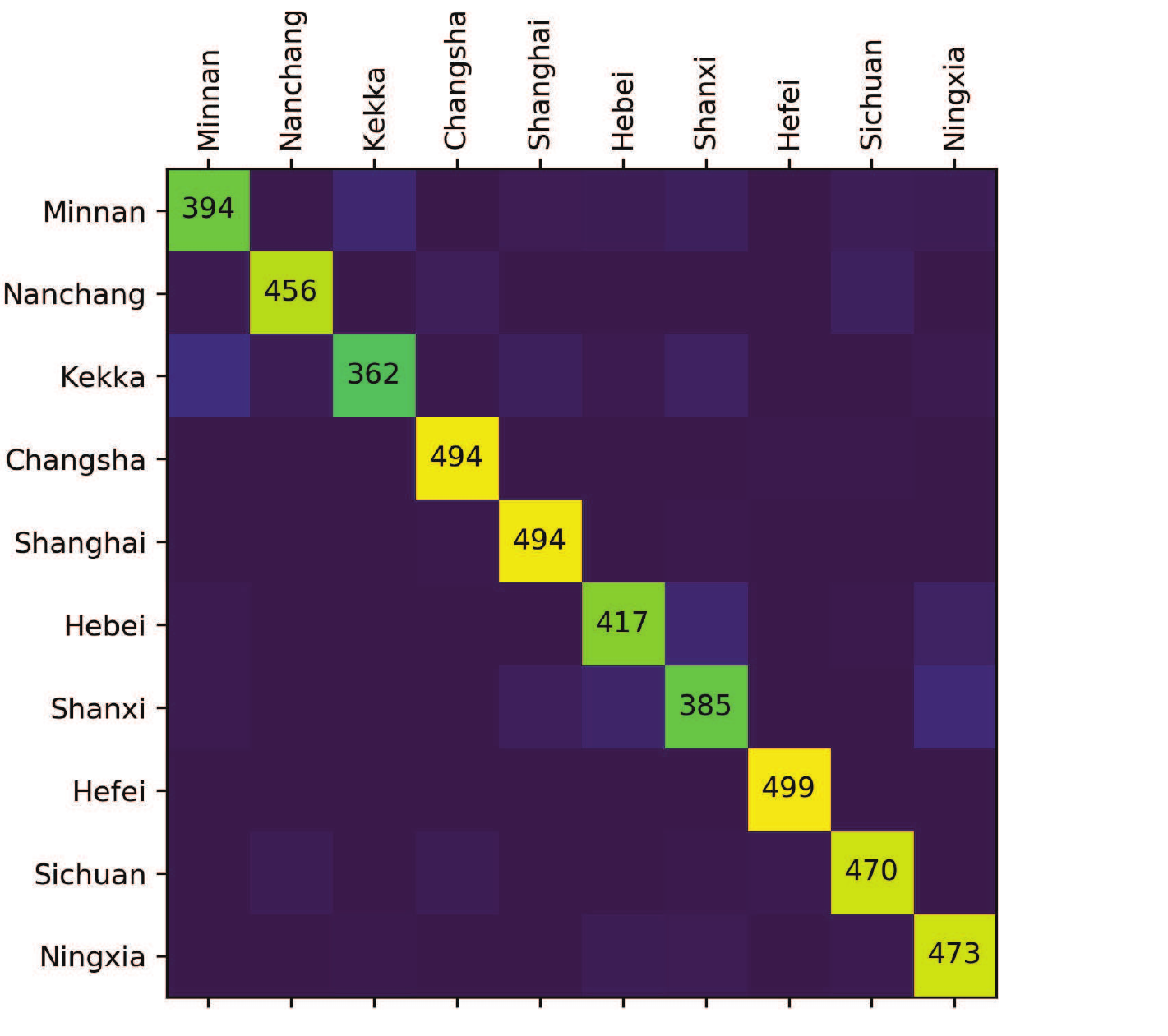}}
 {\includegraphics[scale=0.58]{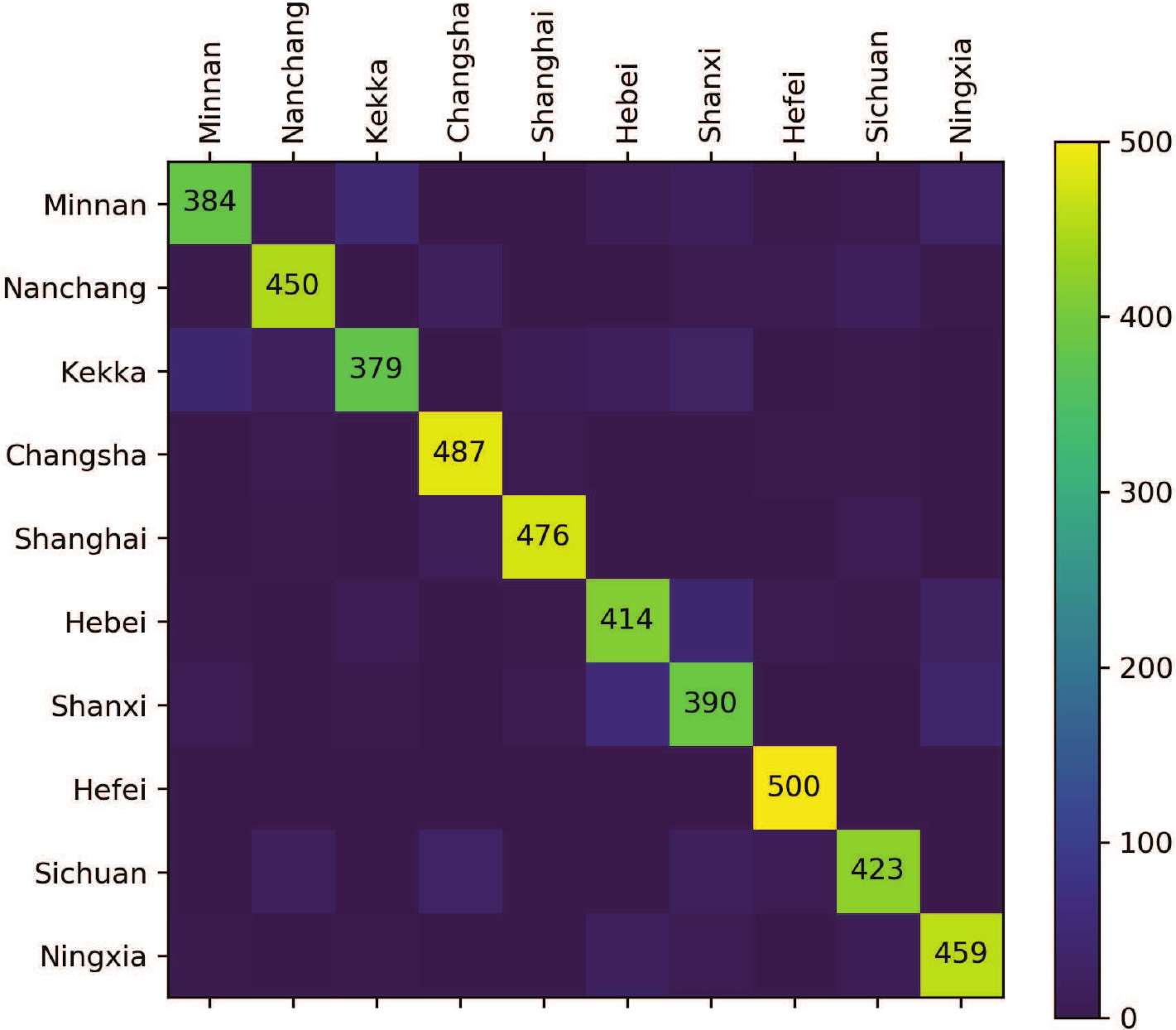}}
\caption{Comparison of confusion matrices produced by the two-stage system (left) and the three-stage system (right)}
\label{fig5}
\end{figure*}

By analyzing the confusing matrices (Figure \ref{fig5}) of predicted results, we can find that the accuracy is high in several dialects' recognition, such as Shanghai (98.8\%) and Hefei (99.8\%), but the systems have some trouble while recognizing Minnan and Kekka, Hebei and Shanxi. The results accord with regional distribution of the dialects. For example, Minnan and Kekka are both in Fujian Province and have lots of cognate words, so it is hard to recognize them in reality.

\begin{table}[!htbp]
\centering
\caption{System Performance of different multi-stage systems}\label{tab2}
\begin{tabular}{ccccc}
\toprule
System Type& RNN& All&$\leq$3s& $>$3s\\
\midrule
Baseline& BLSTM &78.85 &77.60 &80.10 \\
Two-stage& BLSTM &{\bfseries88.88}&{\bfseries87.72} &{\bfseries90.04} \\
Three-stage& BLSTM & 87.24&85.52 &88.96 \\
\bottomrule
\end{tabular}
\end{table}

\subsection{Comparison of different RNN structures}
We further explore the impact of different RNN structures with bidirectional gated recurrent unit (BGRU) and BLSTM. For the two-stage system (Table \ref{tab3}), adding the nodes of BLSTM does not work, but adding another layer makes sense in short-duration task. Moreover, with the same layers and nodes, BLSTM outperforms BGRU in the two sub-tasks. We believe that sound related tasks do not need a very deep network as image related tasks, that is also the reason why we use a shallow ResNet14 as the CNN part.

\begin{table}[!htbp]
\centering
\caption{Comparison of two-stage system with different RNN structures}\label{tab3}
\begin{tabular}{cccccc}
\toprule
RNN& Layers &Nodes& All&$\leq$3s& $>$3s\\
\midrule
BLSTM& 2 &256 &{\bfseries 88.88}& 87.72 &{\bfseries90.04} \\
BLSTM& 2 & 384&88.26&86.96&89.56\\
BGRU& 2 & 256& 88.14&87.16&89.12 \\
BLSTM& 3 & 256 &88.80&{\bfseries87.80} &89.80 \\
\bottomrule
\end{tabular}
\end{table}

We evaluate the three-stage system with the same experiments, and the results (Table \ref{tab4}) demonstrate that the three-stage system can achieve high accuracy in long duration task by larger BLSTM layers and the BGRU structure outperforms BLSTM on the whole.
But adding the third RNN layer also does not work in these experiments.
\begin{table}[!htbp]
\centering
\caption{Comparison of three-stage system with different RNN structures}\label{tab4}
\begin{tabular}{cccccc}
\toprule
RNN& Layers &Nodes& All&$\leq$3s& $>$3s\\
\midrule
BLSTM& 2 &256 &87.24&85.52&88.96 \\
BLSTM& 2 & 384&87.42&85.28&{\bfseries89.56} \\
BLSTM&2 &512&87.52&85.88&89.16\\
BGRU& 2 & 256&{\bfseries88.20}& {\bfseries86.96}&89.44\\
BGRU& 2 & 384& 86.96&85.24&88.68\\
BGRU& 2 & 512&85.74 &83.72 &87.76 \\
BLSTM& 3 & 256 &86.96 &84.88&89.04\\
\bottomrule
\end{tabular}
\end{table}

\begin{table}[!htbp]
\centering
\caption{Epochs to converge of different multi-stage systems}\label{tab5}
\begin{tabular}{cccc}
\toprule
System Type&First stage&Second stage& Third stage\\
\midrule
Baseline& 6 &N/A & N/A\\
Two-stage& 29 &6 & N/A \\
Three-stage& 30 &7 &7\\
\bottomrule
\end{tabular}
\end{table}
As Table \ref{tab5} shows, training networks in the first stage (with CTC loss) needs more time for convergence than training networks in the second or third stage (with cross-entropy loss). We can observe that the two-stage system spends less time while having a slightly higher accuracy compared to the three-stage system.


These two multi-stage systems both much outperform the baseline system. They learn acoustic and language knowledge successively, indicating that language and phoneme are features of different levels, so we have to train step by step to avoid the networks ``forget" some knowledge. 
Through the process, we can find the rules of multi-task and multi-stage training, if the labels are in different levels then multi-stage training should be used such as the situation in our paper, otherwise multi-task training  should be used for parallel learning a wide range of knowledge.

\section{Conclusions}
In this work, we propose an acoustic model based on ResNet14 followed by an RNN to recognize phoneme sequence directly with CTC loss and train a simple RNN lastly to get posteriors for recognizing dialect category, forming a two-stage LID system. The system links the different stages by using intermediate features extracted by a shallow ResNet14 architecture. Compared with a simple network or the three-stage system, the two-stage system achieves the state-of-the-art in the Chinese dialect recognition task. 
We believe this idea of two-stage training can provide inspirations for learning different classes knowledge and can extend to other fields.

\bibliographystyle{IEEEtran}

\bibliography{mybib}

\begin{thebibliography}{10}
\providecommand{\url}[1]{#1}
\csname url@samestyle\endcsname
\providecommand{\newblock}{\relax}
\providecommand{\bibinfo}[2]{#2}
\providecommand{\BIBentrySTDinterwordspacing}{\spaceskip=0pt\relax}
\providecommand{\BIBentryALTinterwordstretchfactor}{4}
\providecommand{\BIBentryALTinterwordspacing}{\spaceskip=\fontdimen2\font plus
\BIBentryALTinterwordstretchfactor\fontdimen3\font minus
  \fontdimen4\font\relax}
\providecommand{\BIBforeignlanguage}[2]{{%
\expandafter\ifx\csname l@#1\endcsname\relax
\typeout{** WARNING: IEEEtran.bst: No hyphenation pattern has been}%
\typeout{** loaded for the language `#1'. Using the pattern for}%
\typeout{** the default language instead.}%
\else
\language=\csname l@#1\endcsname
\fi
#2}}
\providecommand{\BIBdecl}{\relax}
\BIBdecl

\bibitem{richardson2015deep}
F.~Richardson, D.~Reynolds, and N.~Dehak, ``Deep neural network approaches to
  speaker and language recognition,'' \emph{IEEE Signal Processing Letters},
  vol.~22, no.~10, pp. 1671--1675, 2015.

\bibitem{fer2015multilingual}
R.~F{\'e}r, P.~Mat{\v{e}}jka, F.~Gr{\'e}zl, O.~Plchot, and
  J.~{\v{C}}ernock{\`y}, ``Multilingual bottleneck features for language
  recognition,'' in \emph{Sixteenth Annual Conference of the International
  Speech Communication Association}, 2015.

\bibitem{matejka2014neural}
P.~Matejka, L.~Zhang, T.~Ng, H.~S. Mallidi, O.~Glembek, J.~Ma, and B.~Zhang,
  ``Neural network bottleneck features for language identification,'' in
  \emph{Proceedings of Odyssey}, vol. 2014, 2014, pp. 299--304.

\bibitem{jiang2014deep}
B.~Jiang, Y.~Song, S.~Wei, J.-H. Liu, I.~V. McLoughlin, and L.-R. Dai, ``Deep
  bottleneck features for spoken language identification,'' \emph{PloS one},
  vol.~9, no.~7, p. e100795, 2014.

\bibitem{Lopez2014Automatic}
I.~Lopez-Moreno, J.~Gonzalez-Dominguez, O.~Plchot, D.~Martinez,
  J.~Gonzalez-Rodriguez, and P.~Moreno, ``Automatic language identification
  using deep neural networks,'' in \emph{IEEE International Conference on
  Acoustics}, 2014.

\bibitem{lozano2015end}
A.~Lozano-Diez, R.~Zazo-Candil, J.~Gonzalez-Dominguez, D.~T. Toledano, and
  J.~Gonzalez-Rodriguez, ``An end-to-end approach to language identification in
  short utterances using convolutional neural networks,'' in \emph{Sixteenth
  Annual Conference of the International Speech Communication Association},
  2015.

\bibitem{Jin2017End}
M.~Jin, Y.~Song, I.~Mcloughlin, W.~Guo, and L.~R. Dai, ``End-to-end language
  identification using high-order utterance representation with bilinear
  pooling,'' in \emph{Interspeech 2017}, 2017.

\bibitem{garcia2016stacked}
D.~Garcia-Romero and A.~McCree, ``Stacked long-term tdnn for spoken language
  recognition.'' in \emph{INTERSPEECH}, 2016, pp. 3226--3230.

\bibitem{gonzalez2014automatic}
J.~Gonzalez-Dominguez, I.~Lopez-Moreno, H.~Sak, J.~Gonzalez-Rodriguez, and
  P.~J. Moreno, ``Automatic language identification using long short-term
  memory recurrent neural networks,'' in \emph{Fifteenth Annual Conference of
  the International Speech Communication Association}, 2014.

\bibitem{geng2016end}
W.~Geng, W.~Wang, Y.~Zhao, X.~Cai, and B.~Xu, ``End-to-end language
  identification using attention-based recurrent neural networks,''
  \emph{Interspeech 2016}, pp. 2944--2948, 2016.

\bibitem{gelly2017spoken}
G.~Gelly and J.-L. Gauvain, ``Spoken language identification using lstm-based
  angular proximity.'' in \emph{INTERSPEECH}, 2017, pp. 2566--2570.

\bibitem{cai2019utterance}
W.~Cai, D.~Cai, S.~Huang, and M.~Li, ``Utterance-level end-to-end language
  identification using attention-based cnn-blstm,'' \emph{arXiv preprint
  arXiv:1902.07374}, 2019.

\bibitem{lozano2018dnn}
A.~Lozano-Diez, O.~Plchot, P.~Matejka, and J.~Gonzalez-Rodriguez, ``Dnn based
  embeddings for language recognition,'' in \emph{2018 IEEE International
  Conference on Acoustics, Speech and Signal Processing (ICASSP)}.\hskip 1em
  plus 0.5em minus 0.4em\relax IEEE, 2018, pp. 5184--5188.

\bibitem{Tang2017Phonetic}
Z.~Tang, W.~Dong, Y.~Chen, L.~Li, and A.~Abel, ``Phonetic temporal neural model
  for language identification,'' \emph{IEEE/ACM Transactions on Audio Speech \&
  Language Processing}, vol.~26, no.~1, pp. 134--144, 2017.

\bibitem{Masumura2017Parallel}
R.~Masumura, T.~Asami, H.~Masataki, and Y.~Aono, ``Parallel phonetically aware
  dnns and lstm-rnns for frame-by-frame discriminative modeling of spoken
  language identification,'' in \emph{IEEE International Conference on
  Acoustics}, 2017.

\bibitem{Mateju2018}
\BIBentryALTinterwordspacing
L.~Mateju, P.~Cerva, J.~Zdansky, and R.~Safarik, ``Using deep neural networks
  for identification of slavic languages from acoustic signal,'' in \emph{Proc.
  Interspeech 2018}, 2018, pp. 1803--1807. [Online]. Available:
  \url{http://dx.doi.org/10.21437/Interspeech.2018-1165}
\BIBentrySTDinterwordspacing

\bibitem{graves2006connectionist}
A.~Graves, S.~Fern{\'a}ndez, F.~Gomez, and J.~Schmidhuber, ``Connectionist
  temporal classification: labelling unsegmented sequence data with recurrent
  neural networks,'' in \emph{Proceedings of the 23rd international conference
  on Machine learning}.\hskip 1em plus 0.5em minus 0.4em\relax ACM, 2006, pp.
  369--376.

\bibitem{hinton2012deep}
G.~Hinton, L.~Deng, D.~Yu, G.~Dahl, A.-r. Mohamed, N.~Jaitly, A.~Senior,
  V.~Vanhoucke, P.~Nguyen, B.~Kingsbury \emph{et~al.}, ``Deep neural networks
  for acoustic modeling in speech recognition,'' \emph{IEEE Signal processing
  magazine}, vol.~29, 2012.

\bibitem{graves2013speech}
A.~Graves, A.-r. Mohamed, and G.~Hinton, ``Speech recognition with deep
  recurrent neural networks,'' in \emph{2013 IEEE international conference on
  acoustics, speech and signal processing}.\hskip 1em plus 0.5em minus
  0.4em\relax IEEE, 2013, pp. 6645--6649.

\bibitem{graves2014towards}
A.~Graves and N.~Jaitly, ``Towards end-to-end speech recognition with recurrent
  neural networks,'' in \emph{International Conference on Machine Learning},
  2014, pp. 1764--1772.

\bibitem{amodei2016deep}
D.~Amodei, S.~Ananthanarayanan, R.~Anubhai, J.~Bai, E.~Battenberg, C.~Case,
  J.~Casper, B.~Catanzaro, Q.~Cheng, G.~Chen \emph{et~al.}, ``Deep speech 2:
  End-to-end speech recognition in english and mandarin,'' in
  \emph{International conference on machine learning}, 2016, pp. 173--182.

\bibitem{shi2017end}
B.~Shi, X.~Bai, and C.~Yao, ``An end-to-end trainable neural network for
  image-based sequence recognition and its application to scene text
  recognition,'' \emph{IEEE transactions on pattern analysis and machine
  intelligence}, vol.~39, no.~11, pp. 2298--2304, 2017.

\bibitem{zhang2016towards}
Y.~Zhang, M.~Pezeshki, P.~Brakel, S.~Zhang, C.~Laurent, Y.~Bengio, and
  A.~Courville, ``Towards end-to-end speech recognition with deep convolutional
  neural networks,'' \emph{Interspeech 2016}, pp. 410--414, 2016.

\bibitem{wang2017residual}
Y.~Wang, X.~Deng, S.~Pu, and Z.~Huang, ``Residual convolutional ctc networks
  for automatic speech recognition,'' \emph{arXiv preprint arXiv:1702.07793},
  2017.

\bibitem{kim2017residual}
J.~Kim, M.~El-Khamy, and J.~Lee, ``Residual lstm: Design of a deep recurrent
  architecture for distant speech recognition,'' \emph{Proc. Interspeech 2017},
  pp. 1591--1595, 2017.

\bibitem{he2016deep}
K.~He, X.~Zhang, S.~Ren, and J.~Sun, ``Deep residual learning for image
  recognition,'' in \emph{Proceedings of the IEEE conference on computer vision
  and pattern recognition}, 2016, pp. 770--778.

\bibitem{schuster1997bidirectional}
M.~Schuster and K.~K. Paliwal, ``Bidirectional recurrent neural networks,''
  \emph{IEEE Transactions on Signal Processing}, vol.~45, no.~11, pp.
  2673--2681, 1997.

\bibitem{Liu2018}
\BIBentryALTinterwordspacing
Y.~Liu, L.~He, J.~Liu, and M.~T. Johnson, ``Speaker embedding extraction with
  phonetic information,'' in \emph{Proc. Interspeech 2018}, 2018, pp.
  2247--2251. [Online]. Available:
  \url{http://dx.doi.org/10.21437/Interspeech.2018-1226}
\BIBentrySTDinterwordspacing

\bibitem{zhou2018training}
J.~Zhou, T.~Jiang, L.~Li, Q.~Hong, Z.~Wang, and B.~Xia, ``Training multi-task
  adversarial network for extracting noise-robust speaker embedding,''
  \emph{arXiv preprint arXiv:1811.09355}, 2018.

\end{thebibliography}

\end{document}